\documentclass[conference,a4paper]{IEEEtran}

\usepackage[latin1]{inputenc}
\usepackage{subeqnarray, flushend}
\usepackage{amsfonts}
\usepackage{amsmath}
\usepackage{amssymb}
\usepackage{graphicx}
\usepackage{color}
\usepackage{cite}
\usepackage{multirow}
\usepackage[export]{adjustbox}
\usepackage{tablefootnote}
\usepackage{fancyheadings}
\title{Local Radon Descriptors for Image Search}

\author{
	\authorblockN{
		Morteza Babaie\authorrefmark{1}, 
		H.R. Tizhoosh\authorrefmark{2} 
		Amin Khatami\authorrefmark{3} 
		and M.E. Shiri\authorrefmark{1} \vspace{0.1in}
	}
	\authorblockA{
		\authorrefmark{1} Department of Mathematics and Computer Science, Amirkabir University of Technology, Tehran, Iran			
	}
	\authorblockA{
		\authorrefmark{2} KIMIA Lab, University of Waterloo, ON, CANADA
	}
	\authorblockA{
		\authorrefmark{3} Institute for Intelligent Systems Research and Innovation (IISRI), Deakin University, Australia
	}
}

\begin{document}

\maketitle

\begin{abstract}
Radon transform and its inverse operation are important techniques in medical imaging tasks. Recently, there has been renewed interest in Radon transform for applications such as content-based medical image retrieval. However, all studies so far have used Radon transform as a global or quasi-global image descriptor by extracting projections of the whole image or large sub-images. This paper attempts to show that the dense sampling to generate the histogram of local Radon projections has a much higher discrimination capability than the global one. In this paper, we introduce Local Radon Descriptor (LRD) and apply it to the IRMA dataset, which contains 14,410 x-ray images as well as to the  INRIA Holidays dataset with 1,990 images. Our results show significant improvement in retrieval performance by using LRD versus its global version. We also demonstrate that LRD can deliver results comparable to well-established descriptors like LBP and HOG.
\end{abstract}


\section{Motivation}
\label{sec:intro}

Content-Based Image Retrieval (CBIR) refers to a class of algorithms that use the image content (pixels or more commonly their features) to search for similar images in an archive when a query (input) image is given by the user.  Numerous CBIR methods have been introduced that are based on feature descriptors \cite{Nanni20123634,Liu20102380,Kumar2013}. Generally, we can distinguish between feature descriptors that are either based on ``dense sampling'' (features calculated from many small image windows with overlap), or feature descriptors that are based on ``keypoint detection'' (features calculated from small windows around selected points). One may add a third type based on global image features (FFT spectral features, Radon projections etc.) \cite{krig2014computer,Tizhoosh2015}.  The former two types of descriptors, however, have been more widely used and reported in literature. 
  
Local binary patterns (LBP) and the histogram of gradients (HOG) and their extensions are representative for successful methods, which extract local information from images by means of dense sampling. Feature extraction methods such as SIFT and SURF have been commonly used as well.  They are based on keypoint detection scheme which may fail to detect reliable keypoints in medical images \cite{Kashif2016,Sargent2009a}. Dense sampling methods, in contrast, deliver reportedly higher performance \cite{Kashif2016}. 

The LBP generally captures the difference between each pixel and its neighbouring pixels (i.e., 8 pixels) as a binary vector which is then converted to decimal integers that are counted to construct a histogram  \cite{Pietikainen2000}. There are numerous extensions of the LBP by incorporating a different number of neighbors \cite{Shi2007} and different number of histogram bins \cite{Ojala2002}. Uniform LBP reduces the number of bins from 256 to 59 by a selective grouping of patterns. LBP has been successfully applied in various applications such as face detection \cite{Ahonen:2006:FDL:1175897.1176245} and object detection \cite{heikkila2006texture}. It also has been used for medical image retrieval in several reports \cite{Yao2002,Murala2012,Srivastava2017}.

The HOG descriptor, on the other hand, captures the gradient information in densely sampled local windows by counting the number of discrete directions in $n$ bins. It was initially introduced for pedestrian detection \cite{Dalal2005}. However, its power of discrimination made it applicable to a variety of fields in computer vision such as face recognition and image retrieval \cite{Hu2013,deniz2011face}.  


\section{Radon Transfrom}
\begin{figure}[b]
\centering
\includegraphics[width=0.99\columnwidth]{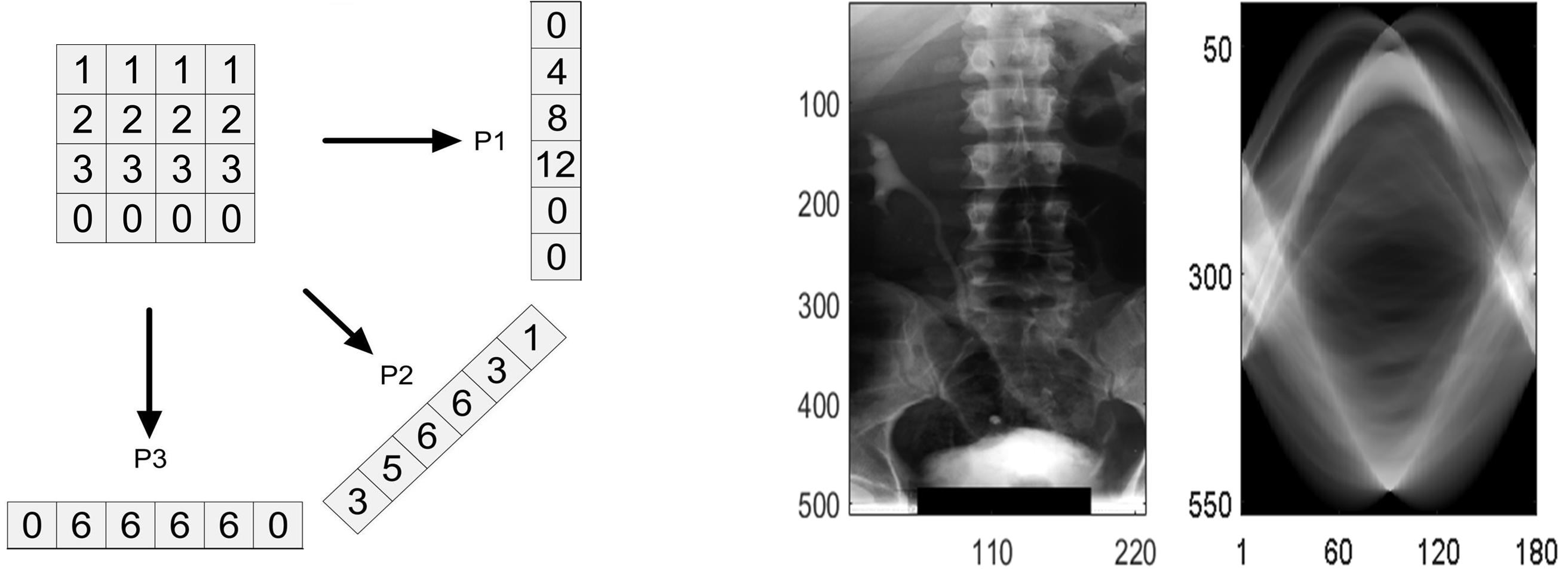}
\caption{Left: Three Radon projections of a $4\times4$ sample window at $0^\circ, 45^\circ$ and $90^\circ$ with zero padding. Right: a sample x-ray image and its sinogram (180 Radon projections).}
\label{fig:Second}
\end{figure}

Radon transform integrates projection values from various directions in order to reconstruct an object (or the human body) from its parallel projections \cite{radon1917determination}. The inverse Radon transform along with filtered backprojection has been used to create body images from projections captured at different directions \cite{Beylkin1985}. Given the spatial intensities $f(x,y)$, the Radon transform can be given as 
\begin{equation}
\mathbf{R}(\rho,\theta)=\int\limits_{-\infty}^{+\infty}\int\limits_{-\infty}^{+\infty} f(x,y)\delta(\rho-xcos\theta-ysin\theta)dxdy,
\label{Radon}
\end{equation}
where $\delta(\cdot)$ is the Dirac delta function. In Figure \ref{fig:Second}, the Radon transform is illustrated for a small window. The Radon transform of a sample image (its \emph{sinogram} containing projections in 180 directions) is depicted as well. 

Several works have recently developed approaches to medical image retrieval using ``Radon barcodes'' in which they binarize a certain number of equidistant projections \cite{Tizhoosh2015,Tizhoosh2016,Zhu2016}. We find the idea of using \emph{projections} quite intriguing, specially for the medical field. However, the literature only contains projections-based descriptors that use ``global'' information (Radon projections of the entire image). Of course, the performance of such methods, restricted to global information, is expectedly rather limited.

There have been several works that use the term ``local Radon transform'' in connection with grayscale images (and not binary ones). However, they generally mean ``localized'' application of Radon transform on a certain part of the image or a sub-image \cite{krause2010micro,pourreza2008radon}. 

In this paper, we introduce a new dense-sampling descriptor based on Radon transform named \emph{Local Radon Descriptor} (LRD). We describe the details of LRD in the next section. We use two image datasets to validate the proposed approach against LBP and HOG although our primary agenda is to improve methods which have used ``global'' Radon projections to assemble compact descriptors. LRD is supposed to break through the limitations of globally captured projections by examining gradient change of locally extracted projections.

\section{Local Radon Descriptor (LRD)}
As mentioned before, several approaches have been proposed to use Radon projections and/or their binary versions for retrieval of medical images. All these methods so far only employ ``global'' projections; they apply Radon transform on the entire image. However, we know that a major reason for the success of LBP and HOG is their capability of capturing ``local'' information in form of a histogram. Combining this knowledge with what Radon projections can quantify, it is obvious that one has to design a new class of projection-based descriptors that can exploit local gradient information.  

\vspace{0.1in}
To construct the LRD, we go through the following steps:


\textbf{Step 1.} For every selected image block (local window), $n$ equidistant Radon projections are calculated. We specify each element in each projection by $\mathbf{R}(\rho_{i},\theta_{j})$, where 

\begin{equation}
\theta_{j}=\left\{0,\dfrac{\pi}{n},\dfrac{2\pi}{n},\dots,\dfrac{n\pi}{n}-1\right\},
\end{equation} 

and  $\rho_i$ refers to the $i$-th line of the projection under the $j$-th direction.

\textbf{Step 2.} For all projections, we build the first order derivative vector $\mathbf{d}_{\theta_j}$ to quantify projection change:
\begin{equation}
\mathbf{d}_{\theta_j}(i)=\frac{\partial}{\partial \rho} \mathbf{R} \left(\rho,\theta_{j}\right)=\mathbf{R}(\rho_{i+1},\theta_{j})-\mathbf{R}(\rho_{i},\theta_{j}).
\label{DVs}
\end{equation}


\textbf{Step 3.} Looking at each projection $\mathbf{p}_i=\mathbf{R}(\rho_{k=1:L},\theta_{j})$ of length $L$, we build $n/2$ pairs of projections $(\mathbf{p}_{\theta_i},\mathbf{p}_{\theta_j})$ whereas $\theta_j=\theta_i+\alpha$.

\textbf{Step 4.} Selecting two derivative vectors $\mathbf{d}_{\theta}$ and $\mathbf{d}_{\theta+\alpha}$, a ratio vector $\mathbf{r}$ is created by dividing the elements of both gradient vectors resulting in a vector in $(-\infty, +\infty)$:
\begin{equation}
\mathbf{r}(i)=\frac{\mathbf{d}_{\theta}(i)}{\mathbf{d}_{\theta+\alpha}(i)}.
\label{RV}
\end{equation}
To quantize each element of $\mathbf{r}$ and count its occurrence, we apply the \emph{arctan} operator to map the $\mathbf{r}$ elements from $(-\infty, +\infty)$ to  $[-\pi, +\pi]$. Then, we can discretize the outcome of the $\tan^{-1}(\mathbf{r})$ in $b$ different bins resulting in a vector $\mathbf{r}_q$. 

\textbf{Step 5.} We also build a summation vector $\mathbf{s}$ by adding the absolute values of the derivatives:
\begin{equation}
\mathbf{s}(i)=|\mathbf{d}_{\theta}(i)+\mathbf{d}_{\theta+\alpha}(i)|
\label{SV}
\end{equation}

\textbf{Step 6.} We capture the information of two Radon signals in the $k$-th local histogram $\mathbf{h}_k(i)$ by counting the quantized $\mathbf{r}_q$ elements and weighting them by corresponding $\mathbf{s}$ elements. After generating all histograms, the main descriptor is created by concatenating individual histograms $\mathbf{h}_k$. The length of the descriptor is proportional to the number of blocks per column $n_\textrm{column}$, the number of blocks per row $n_\textrm{rows}$, and the number of bins $b$ per histogram in the blocks: length$=n_\textrm{column}\times n_\textrm{rows}\times b$.

\begin{figure*}[bt]
\centering
\includegraphics[width=\textwidth]{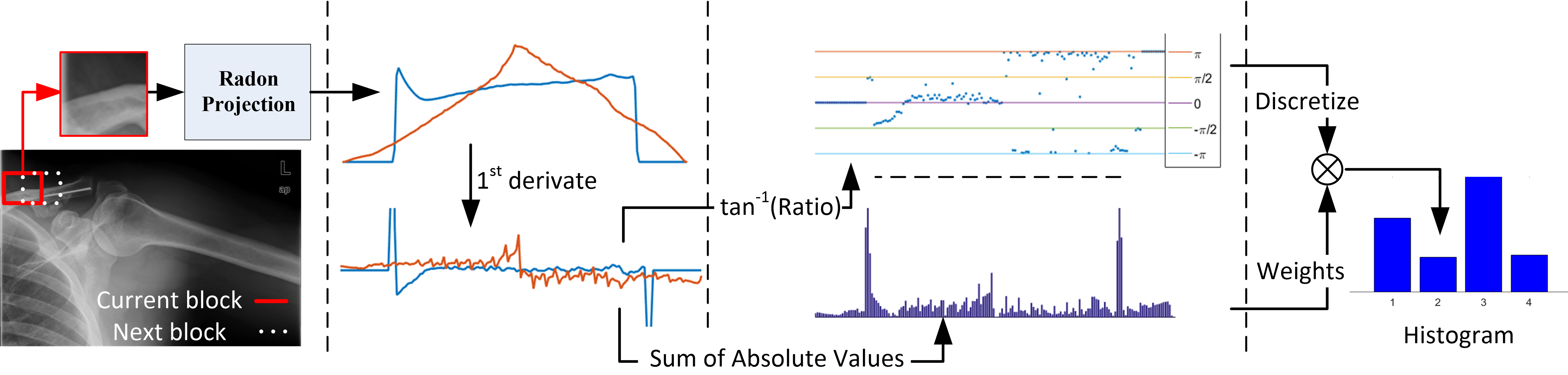}
\caption{General steps to extract LRD from gradients of pairs of Radon projections. The arctan of the ratio and the sum of absolute values of paired projections are used to assemble a histogram for each sub-image.}
\label{fig:First}
\end{figure*}

\vspace{0.1in}
Figure \ref{fig:First} shows the steps for creating a histogram $\mathbf{h}_k$ for a sub-image. 

\section{Experiments}
\label{sec:results}
We conducted two different experiments to evaluate the LRD versus the global Radon descriptors. The first experiment was conducted on the IRMA dataset, and the second one uses the INRIA Holidays dataset. In the following, we briefly describe these datasets  and then the results of LRD for retrieving images from these datasets will be reported. The Matlab code for LRD is available on our website\footnote{\emph{http://kimia.uwaterloo.ca/}}.

\subsection{Datasets}

\begin{figure}[tb]
  \centering
  \centerline{\includegraphics[width=7.5cm]{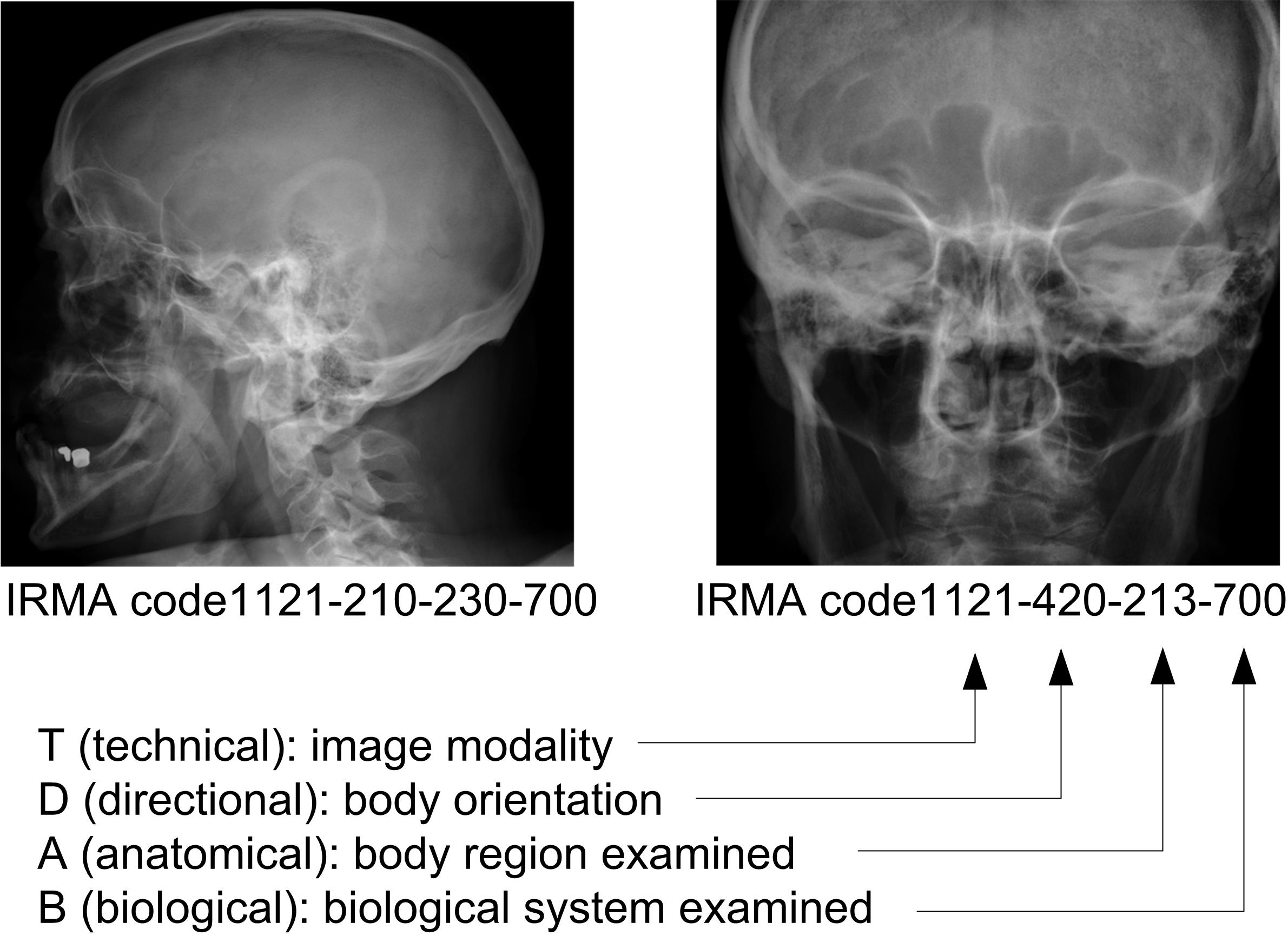}}
\caption{Two sample x-ray images and their corresponding IRMA codes.}
\label{fig:Iramacode}
\end{figure}

\textbf{IRMA dataset --} The Image Retrieval in Medical Applications (IRMA) dataset, created by the Aachen University of Technology, has been used to evaluate retrieval methods for medical CBIR tasks \cite{Tommasi2009}. It contains 12,677 images in the training set and 1,733 images in the testing set. The IRMA dataset is extremely challenging mainly due to its highly imbalanced distribution of 193 different classes. As well, digitized landmarks and annotations make the retrieval  even more difficult. The IRMA x-ray images come with a unified code to calculate the retrieval error. As shown in Figure \ref{fig:Iramacode}, all images in the dataset have been assigned a 13-digit \emph{IRMA code}. The error per query image is a number between $0$ and $1$. As a result, the total IRMA error for all test images is a number between 0 and 1,733. We used the Python code provided by ImageCLEFmed09 to compute the errors\footnote{\emph{http://www.imageclef.org/2008/medical}}. The dataset can be downloaded from IRMA website\footnote{\emph{https://ganymed.imib.rwth-aachen.de/irma}}.

\begin{figure*}[htb]
  \centering
  \centerline{\includegraphics[width=0.65\textwidth]{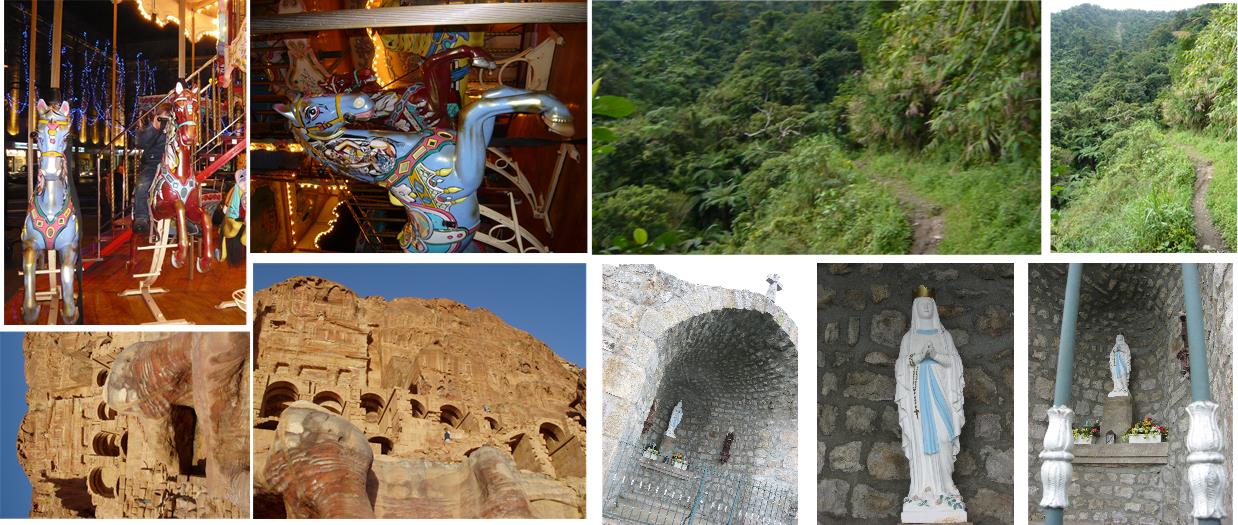}}
\caption{Sample images from the Holidays dataset.}
\label{fig:Holy}
\end{figure*}

\textbf{INRIA Holidays dataset --} Jegou et al. \cite{Jegou2008} have created a searching dataset containing 1,991 high resolution images in 500 different categories such as landscapes, buildings and historical sites from different point of views and with various sizes (Figure \ref{fig:Holy}). The first image in each category is supposed to be used as a query image. The performance of retrieving method is evaluated via the percentage of true retrieval against all 500 query images. The dataset can be downloaded from INRIA website\footnote{\emph{http://lear.inrialpes.fr/~jegou/data.php}}.

\subsection{Parameter Setting} 
There are several parameters which influence the performance of LRD and hence must be investigated. Perhaps two important factors to be adjusted are the number of blocks per row, $n_\textrm{rows}$, and per column, $n_\textrm{columns}$. Note that these two parameters are needed to investigate in LBP and HoG as well.  Additionally, the overlap percentage $O_p$ between adjacent blocks is certainly also a factor that may affect the retrieval performance. And finally, we have to also specify how many projection pairs should be selected, and which projections should be paired to achieve the best results. 

Through zero-padding, all images are resized to $256\times 256$ (hence, $n_\textrm{columns}=n_\textrm{rows}$). We tested 2 different ways to adjust pair selection among $n=18$ projections:

\begin{enumerate}
\item \textbf{Orthogonal selection:} With $n=18$, we have $\theta = \{ 0^\circ,10^\circ,20^\circ,\dots,170^\circ\}$. Orthogonal angles are then selected as a pair such that we have a set $P$ of 9 paired projections: 
\begin{equation}
P=\{(0^\circ,90^\circ),(10^\circ,100^\circ),\dots,(80^\circ, 170^\circ)\}.
\end{equation}
  
\item \textbf{Using characteristic projections:} Based on \cite{Babaie2017}, $\theta=0^\circ$ and $90^\circ$ are the most characteristic projection directions to retrieve images from the IRMA dataset (x-ray images generally do not exhibit rotations, and these directions provide very distinct projections). In order to exploit this observation, we paired each projection under $\pi /2$ by zero degree and all other projections by 90 degree such that we get a set $P=P_{0^\circ}\cup P_{90^\circ}$ with
\begin{equation}
P_{0^\circ}=\{( 0^\circ,10^\circ),(0^\circ,20^\circ),\dots, (0^\circ,80^\circ)\},
\end{equation}
and 
\begin{equation}
P_{90^\circ}=\{( 90^\circ,100^\circ),(90^\circ,110^\circ),\dots, (90^\circ,170^\circ)\}.
\end{equation}
 \end{enumerate}
 

 
 The number of blocks and bins were tuned by exhaustive search. The best number of blocks for INRIA Holidays dataset was set to $n_\textrm{columns}=n_\textrm{rows}=3$ (9 sub-images) and the best number of bins was determined to be $b=22$. For IRMA dataset, these number were set to $n_\textrm{columns}=n_\textrm{rows}=5$ (25 sub-images), and $b=12$, respectively. The orthogonal paring did not exhibit good results such that we conducted all experiments with the characteristic projection paring $P_{0^\circ}$ and $P_{90^\circ}$.

\begin{figure}[tb]
\centering
\includegraphics[width=4cm]{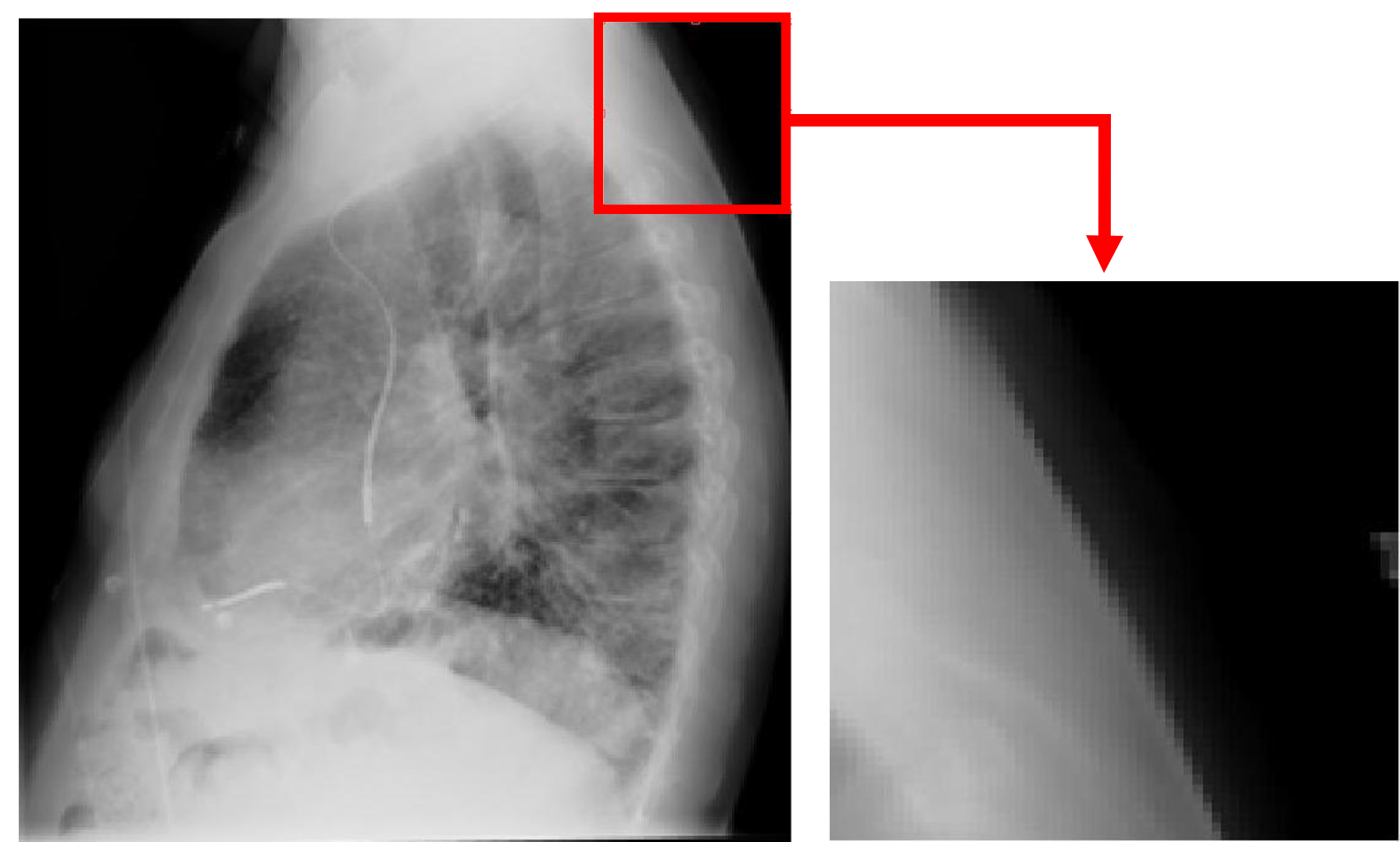}\\
\includegraphics[width=0.32\textwidth]{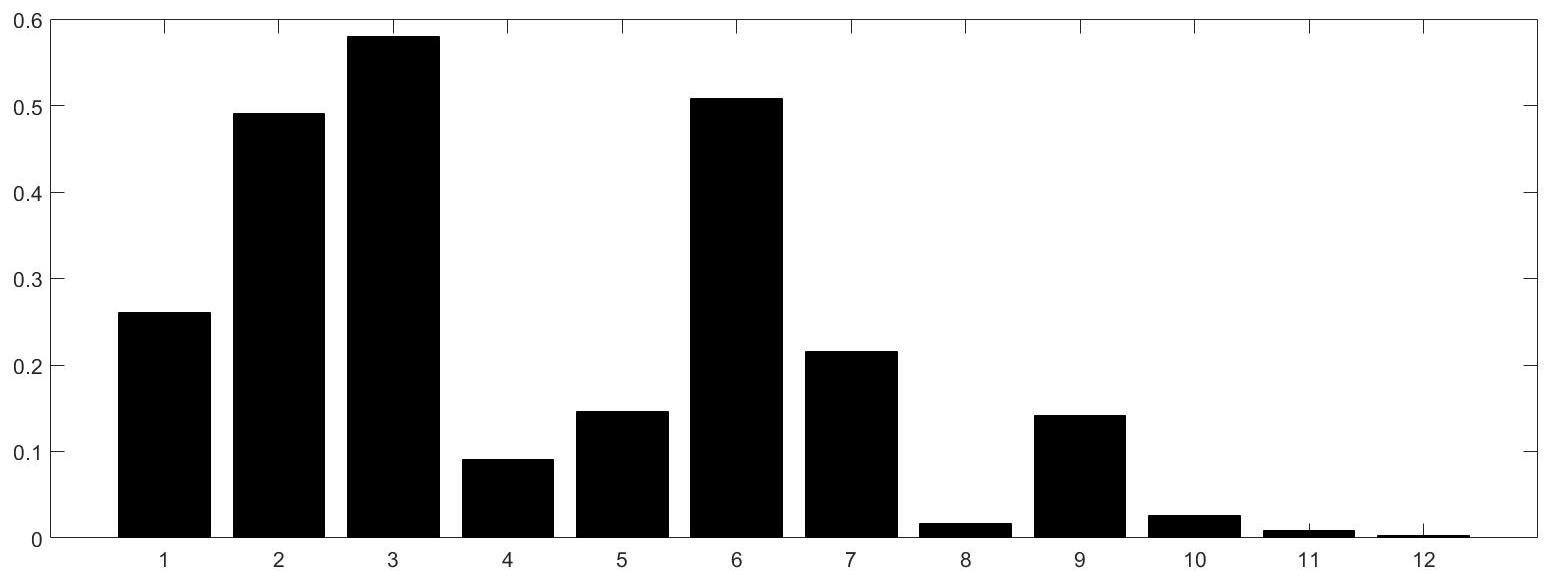}
\includegraphics[width=0.32\textwidth]{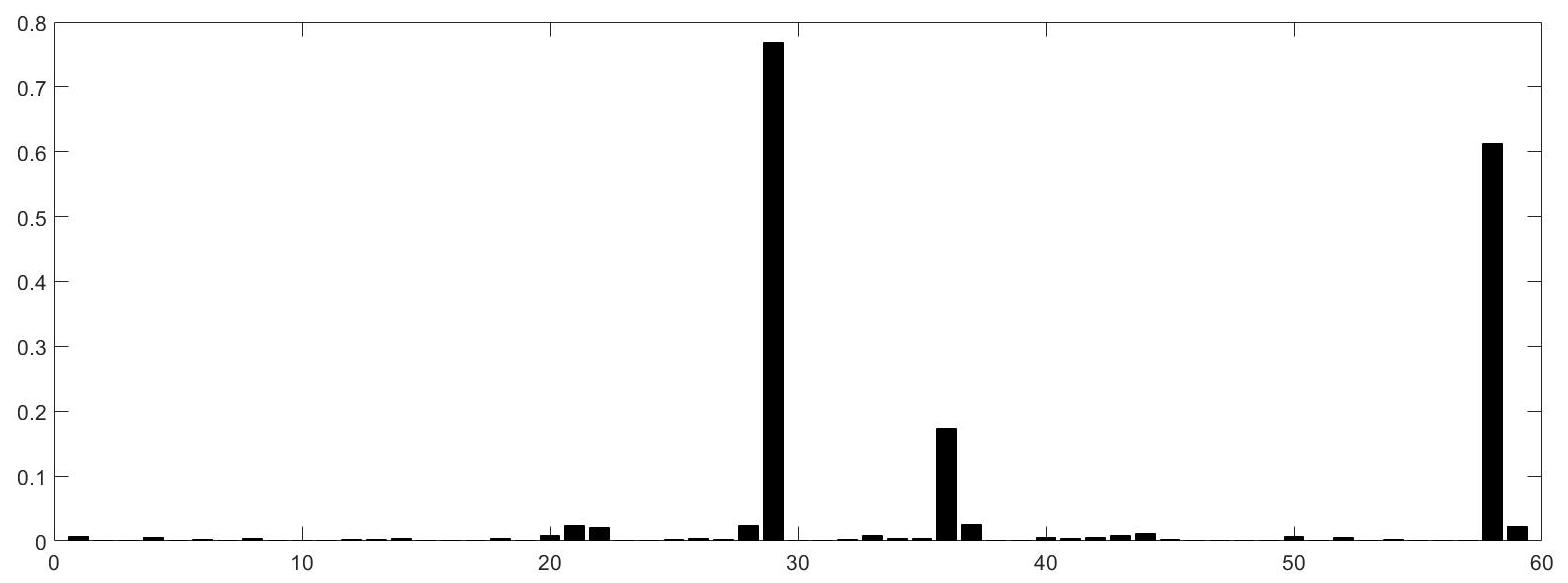}
\includegraphics[width=0.32\textwidth]{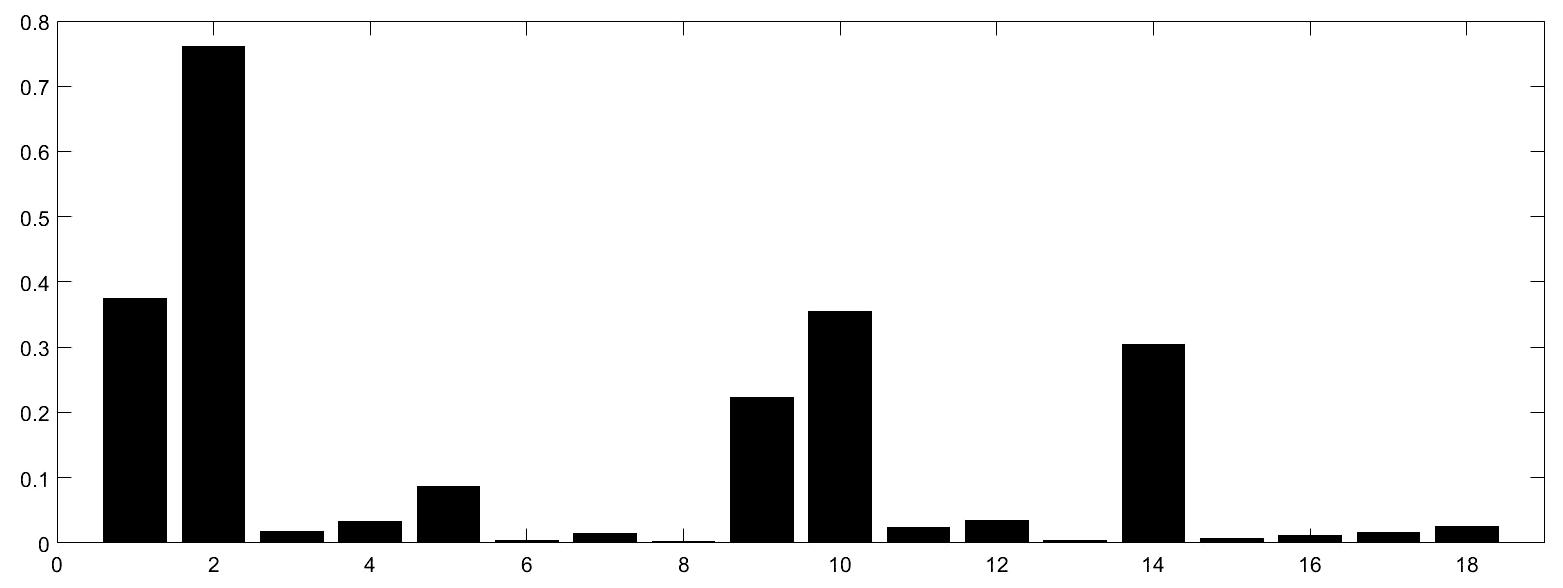}
\caption{Top: A sample x-ray and one of its sub-images. Bottom (left to right): LRD, LBP and HOG histograms of the sub-image.}
\label{fig:samplehist}
\end{figure}

\vspace{0.1in}
Fig. \ref{fig:samplehist} shows sample histograms for LRD in comparison with LBP and HOG. Table \ref{tab:details} lists the histogram length ($=n_\textrm{columns}\times n_\textrm{rows}\times b$) for all descriptors for the two datasets. 

\begin{table}[htp]
\caption{Sub-image and bin settings for LBP, HOG and LRD ($n_\textrm{columns}\times n_\textrm{rows}\times b$).}
\begin{center}
\begin{tabular}{l|l|l}
& IRMA dataset & Holidays dataset \\ \hline
LBP & $12\times 12\times 59$ & $5\times 5\times 33$ \\
HOG & $14\times 14\times 18$ & $3\times 3\times 24$ \\
LRD & $5\times 5\times 12$ & $3\times 3\times 22$ \\ \hline
\end{tabular}
\end{center}
\label{tab:details}
\end{table}%

\subsection{Results}
For sake of comparison, we used well-established descriptors like HOG and LBP to retrieve images as well. We did the same parameter tuning steps for number of blocks and the number of bins for HOG and the cell size in LBP. The results of applying all three methods on the IRMA dataset is compared with the best global Radon (GR) results as reported in literature \cite{Babaie2017}. The maximum error is 1733 (if the retrieval for all test images delivers wrong macthes). Hence, beside IRMA error, we can also report the accuracy ($=1-\frac{\textrm{error}}{1733}$).

\begin{figure}[tb]
\centering
\includegraphics[width=3cm,height=1.2cm,fbox]{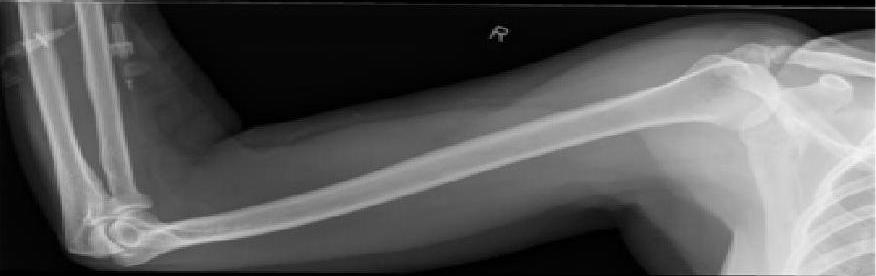}
\includegraphics[width=3cm,height=1.2cm]{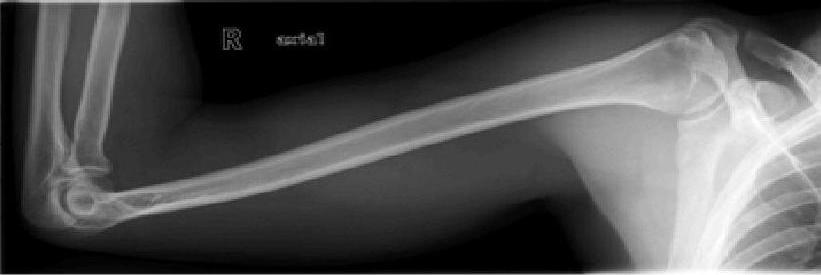}
\includegraphics[width=3cm,height=1.2cm]{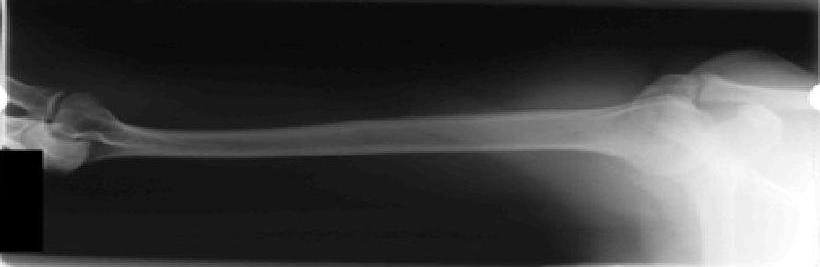}
\includegraphics[width=3cm,height=1.2cm]{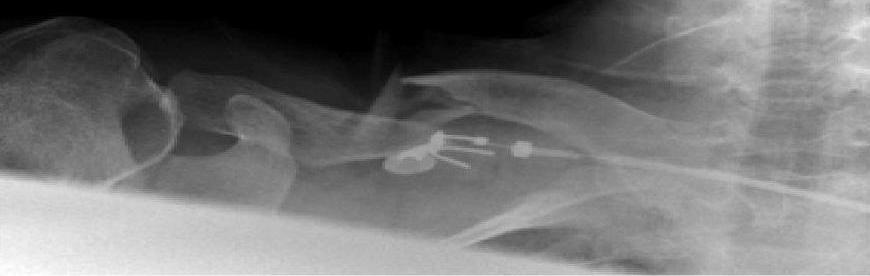}\\
\includegraphics[width=3cm,height=1.2cm,fbox]{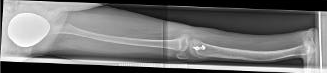}
\includegraphics[width=3cm,height=1.2cm]{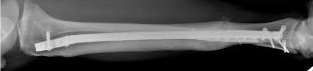}
\includegraphics[width=3cm,height=1.2cm]{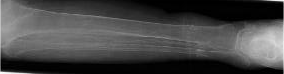}
\includegraphics[width=3cm,height=1.2cm]{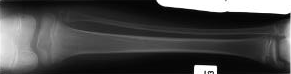}
\caption{Sample retrieval results for LRD (left image is the query image). The top three results are shown. Top: The first two results are correct (IRMA errors: 0, 0, and 0.58), bottom: the first results is wrong (IRMA errors: 0.24, 0.45 and 0.45).}
\label{fig:retresults}
\end{figure}

Fig. \ref{fig:retresults} shows sample retrieval results for LRD. We display the top three matches for query images whereas for error calculation we only consider the first match. 

Table \ref{results table}, shows the comparison between these four retrieving methods (LRD, global Radon projections, HOG and LBP) as well as their descriptor lengths. The  retrieval times via $k$-NN (with $k=1$) are provided as well. LRD performance is slightly below the LBP and close to HOG. However, we can observe significant improvement in local implementation of Radon projections versus global methods. The IRMA error is considerably reduced (accuracy increase of more than 10\%) and the LRD descriptor length is 4 times shorter than its global counterpart. 

Although LBP and HOG are more accurate for the IRMA images, their descriptor lengths are 28 and 12 times, respectively, longer than LRD's, a fact that is also reflected in computation times. Another interesting point is, for the INRIA Holidays dataset, that LRD results are the same as HOG's. Also for this dataset, LRD delivers a shorter descriptor.

\begin{table*}[tb]
\centering
\caption{Retrieval results (GR stands for global Radon). LRD considerably improves the performance of projection-based descriptors as proposed via global methods. As well, LRD does in fact deliver comparable results with state-of-the-art descriptors like LBP and HOG.}
\label{results table}
\begin{tabular}{|l|l|l|l|l|l|l|}
\hline
\multicolumn{1}{|c|}{\multirow{2}{*}{Method}} & \multicolumn{3}{c|}{IRMA Dataset}  & \multicolumn{3}{c|}{Holidays Dataset} \\ \cline{2-7}
\multicolumn{1}{|c|}{}                        & Error (Accuracy) & Length & Time   & True    & Length   & Time  \\ \hline
LBP                                           & 244.81 (85.87\%)   & 8496   & 0.50s & 48.00\%   & 825      & 0.03s  \\ \hline
HOG                                          & 266.36 (84.46\%)  & 3528   & 0.23s & 40.00\%    & 216      & 0.0s  \\ \hline
LRD                                           & 287.77 (83.39\%)  & 300    & 0.02s & 40.02\%   & 198      & 0.0s  \\ \hline
GR                                  & 472.31 (72.74\%)   & 1200   & 0.11s & 06.00\%   & 2400     & 0.1s  \\ \hline
\end{tabular}
\end{table*}

%


\section{Summary}
Radon projections can be used to construct descriptors for image retrieval. However, global projections (projecting the entire image) may not deliver expressive features suitable for image retrieval. In this paper, we introduced Local Radon Descriptor (LRD) as a new descriptor for image retrieval. It is designed to capture Radon projections in local neighbourhoods by looking at the ratio and sum of derivatives of paired projections. LRD exhibits significant advantage over reported works on projection-based approaches to image retrieval that are globally implemented. As well, LRD is a very compact descriptor. Experimental results on two datasets demonstrated the potentials of LRD for both medical and non-medical images. 

\bibliographystyle{IEEEtran}
\bibliography{references}

\end{document}